\documentclass[10pt,twocolumn,letterpaper]{article}

\usepackage{cvpr}              

\usepackage{graphicx}
\usepackage{amsmath}
\usepackage{amssymb}
\usepackage{booktabs}
\usepackage{bbding}
\usepackage{multirow}
\usepackage{subcaption}
\usepackage{tabularx}

\usepackage[pagebackref,breaklinks,colorlinks]{hyperref}
\usepackage[dvipsnames]{xcolor}
\usepackage{colortbl}
\usepackage{balance}

\usepackage[capitalize]{cleveref}
\crefname{section}{Sec.}{Secs.}
\Crefname{section}{Section}{Sections}
\Crefname{table}{Table}{Tables}
\crefname{table}{Tab.}{Tabs.}

\newcommand{\yty}[1]{\textcolor{black}{#1}}
\newcommand{\ytyn}[1]{\textcolor{black}{#1}}

\newcommand{\CUT}[1]{}
\definecolor{darkergreen}{RGB}{21, 152, 56}

\begin{document}
	
\title{SimVTP: Simple Video Text Pre-training with Masked Autoencoders}
	
\author{Yue Ma$^{1\dagger}$ \quad
Tianyu Yang$^{2\ddagger}$ \quad
Ying Shan$^2$ \quad
Xiu Li$^{1\ddagger}$ \\
$^1$Tsinghua Shenzhen International Graduate School, Tsinghua University \quad 
$^2$Tencent AI Lab
}
\maketitle

\newcommand\blfootnote[1]{%
  \begingroup
  \renewcommand\thefootnote{}\footnote{#1}%
  \addtocounter{footnote}{-1}%
  \endgroup
}

\blfootnote{$\dagger$ Work done during an internship at Tencent AI Lab.}
\blfootnote{$\ddagger$ Corresponding author.}

\begin{abstract}
	
	\yty{This paper presents SimVTP: a \textbf{Sim}ple \textbf{V}ideo-\textbf{T}ext \textbf{P}re-training framework via masked autoencoders. We randomly mask out the spatial-temporal tubes of input video and the word tokens of input text and then feed them into a unified autencoder to reconstruct the missing pixels and words.  Our SimVTP has several properties: 1) Thanks to the unified autoencoder,  SimVTP reconstructs the masked signal of one modality with the help from another modality, which implicitly learns the cross-modal alignment between video tubes and text tokens. 2) SimVTP not only benefits from a high video masking ratio (e.g. 90\%) due to the temporal redundancy of video, but also needs a high text masking ratio (e.g. 75\%), which is much higher than BERT (e.g. 15\%), to achieve optimal performance.  This is because the aid of video modality makes text reconstruction less challenging, which thus needs a higher mask ratio to make the pretext harder for useful feature learning. 3) Equipping SimVTP with video-text contrastive learning (VTC) and video-text matching (VTM), which are two commonly used cross-modal training strategies, could further improve the transferable performance significantly. 4) SimVTP is data-efficent, e.g., pre-training only on 10\% data of WebVid-2M, SimVTP achieves surprisingly good results (43.8 R@1) on MSRVTT, which is far above recent state-of-the-art methods pre-trained on both CC3M and WebVid-2M.  We transfer our pre-trained model to various downstream tasks and achieve superior performance. The codes and models will be released at \href{https://github.com/mayuelala/SimVTP}{https://github.com/mayuelala/SimVTP}.
 }
	
\end{abstract}

\section{Introduction}
\label{sec:intro}

\begin{figure}
	\includegraphics[width=8cm]{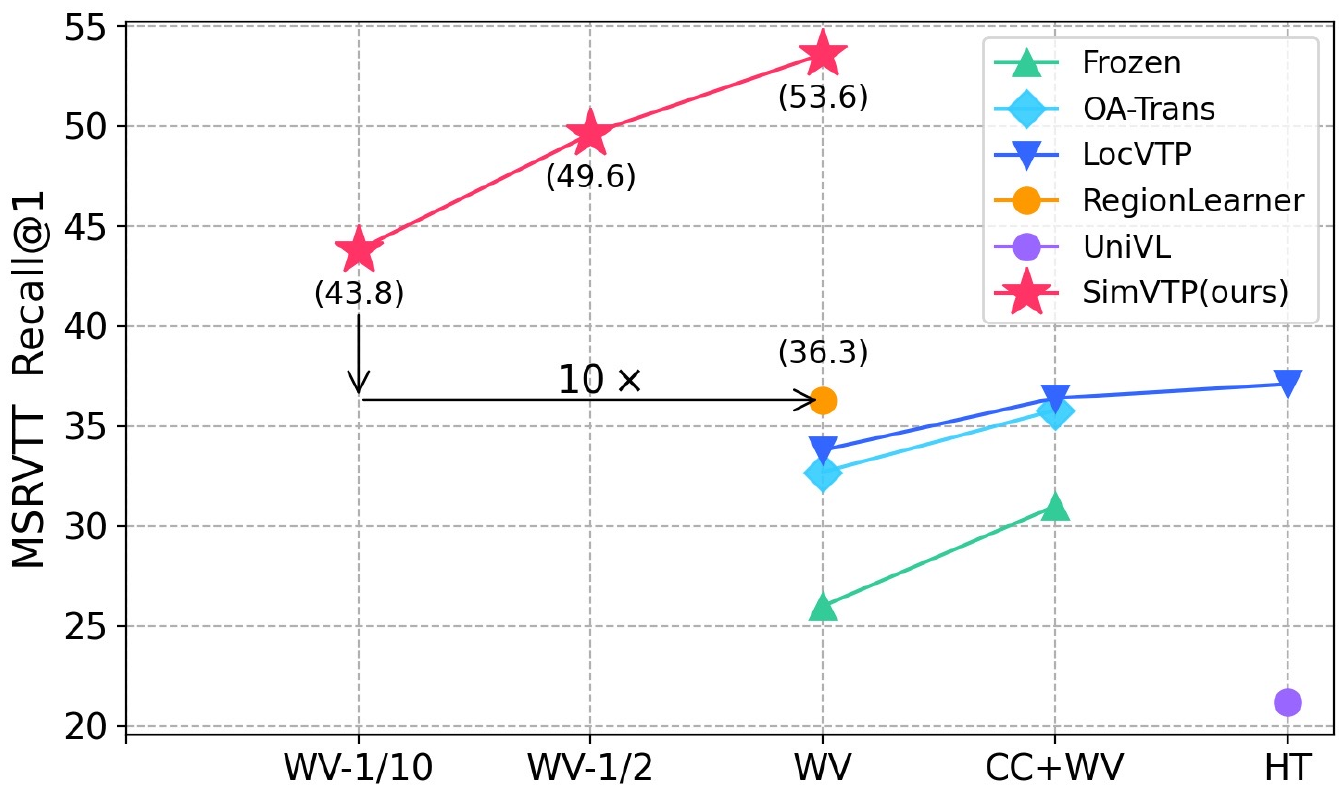}
	\caption{\textbf{SimVTP is data-efficient. }With only 1/10 of WebVid-2M as training data, SimVTP achieves 43.8\% R@1 on MSRVTT which outperforms most recent state-of-the-art methods.}
	\label{fig:1}
	\vspace{-0.5cm}
\end{figure}

\yty{Video-text pre-training has attracted surging attention due to its capability of learning transferable multi-modal representations for a broad range of downstream tasks, e.g., text-video retrieval\cite{cheng2021improving, bain2021frozen, yan2021video, wang2022object, liu2022ts2net}, video question answering\cite{fu2021violet, yang2021just, yu2018joint, torabi2016learning}, video grounding\cite{patrick2020support, miech2020end, tang2021decembert}, etc.. Pioneering works \cite{wang2022object, li2022align} have demonstrate promising results by pre-training on large scale video-text datasets, like WebVid-2M \cite{bain2021frozen} and Howto100M \cite{miech2019howto100m}. Among them, the mainstream methods \cite{bain2021frozen, cao2022locvtp} learn the joint video-text representation in a contrastive framework by pushing paired video-text close and repelling unpaired ones apart.  To further enhance the feature learning, OA-Trans \cite{wang2022object} proposes to align the fine-grained semantic contents of video and text  by using an object detector to detect objects in advance.   RegionLearner \cite{yan2021video} designs a
	model which can take into account the structure of objects during pre-training on large-scale video-text pairs, These designs complicate the pre-training framework which thus weaken its scalability. }
\yty{Inspired by recent success of masked autoencoders in NLP and vision pre-training,  we propose a simple video-text pre-training framework, which is dubbed as SimVTP, by randomly masking out video tubes and text tokens and then reconstructing them.  Not only does SimVTP greatly outperform the mainstream video-text pre-training methods, but also it is simpler, requiring neither object detector \cite{ren2015faster, cai2019cascade} nor a region learner \cite{yan2021video}.}

\yty{SimVTP has several interesting properties: 
	Firstly, due to the use of a unified transformer to encode both video tubes and text tokens in SimVTP,  one modality is reconstructed with the aid of the other one. This implicitly learns the tube-word alignment between paired video text sample, which has been demonstrated to be effective for good feature learning \cite{li2022align, li2021align}.  This jointly learned unified encoder (i.e. a Transformer) can be used to extract both video and text representation separately for downstream tasks. 
	Secondly, it has been shown that video is information-sparse and has heavy spatial temporal redundancy \cite{tong2022videomae}, which therefore needs an extremely high mask ratio (e.g. 90\%) to obtain optimal performance.  Conventionally, texts are masked with a lower mask ratio (e.g. 15\% in BERT \cite{devlin2018bert}) since languages are highly semantic and information-dense \cite{he2022masked}. Masking out too much words would make the pre-training task so challenging that it hard to learn good features. Interestingly, we observe that SimVTP benefits from high mask ratios on both video and text, e.g.,  achieving best performance with 90\% and 75\% mask ratio on video and text respectively. We hypothesize that this is because that text reconstruction becomes less challenging since the coupled video, which conveys the same information with text,  provides extra information redundancy. 
	Thirdly, SimVTP can be seamlessly armed with video-text contrastive learning (VTC) and video-text matching (VTM), which are two commonly used cross-modal training strategies, and be further improved with a large margin. We find that randomly masking is a robust data augmentation method that works surprising well in contrastive frameworks, which is in line with \cite{jing2022masked}. VTC with masked video-text as input significantly outperforms the one with original data pair as input, while also saving a large amount of compute and GPU memory since the video encoder only operates on visible tubes which is a small portion (10\%) of all tokens.
	Finally, SimVTP is a data-efficient learner. With only 10\% data (240K) of WebVid-2M for pre-training, SimVTP achieves 43.8\% R@1 on MSRVTT, which significantly outperforms recent advanced methods\cite{bain2021frozen, wang2022object, cao2022locvtp} pre-trained on much larger datasets, e.g., CC3M (3M), WebVid-2M (2.5M) and Howto100M(100M) (Figure \ref{fig:1}). With moderate data scale and model size, SimVTP achieves new state-of-the-art results on a broad range of downstream tasks. For instance, when using a ViT-Base pre-training on WebVid-2M,  SimVTP obtains 53.6\% R@1 on MSRVTT (text to video retrieval), 44.7\% accuracy on MSRVTT-VQA (video question answering) and 49.2\%$R_1^{0.5}$ on ActivityNet (video grounding)}.

Our main contributions are summarized as three-fold:
\begin{itemize}
	\item 
	We propose a simple \yty{yet} effective video-text pre-training framework with masked autoencoders,  dubbed as SimVTP, which \yty{jointly encodes video and text in a unified framework to reconstruct missing video tubes and words.}
	
	\item 
	\yty{We find that utilizing the \textit{randomly masked }video-text samples to perform video-text contrastive learning (VTC) and video-text matching (VTM), which are two commonly used cross-modal training strategies, could further improve the transferable performance significantly. }
	
	\item 
	\yty{We conduct extensive experiments on three downstream video-text tasks, e.g., video text retrieval, video question answering and video grounding and achieves significant improvement compared with recent state-of-the-art methods. }

\end{itemize}

\section{Related work}
\label{sec:Related}
\noindent\textbf{Video-Text Pre-training.}
\yty{Given large-scale paired video-text data, e.g. WebVid-2M \cite{bain2021frozen} and HowTo100M \cite{miech2019howto100m}, video-text pre-training is the task of learning a joint cross-modal representation to support a broad scope of downstream tasks. }
In general,
the mainstream methods could be divided into two categories:
(1) Generative methods:
\yty{Those }approaches \cite{li2019visualbert,fu2021violet,tang2019coin,liu2022ts2net, chen2020uniter, wang2021dig} attempt to extend BERT-like\cite{zhu2020actbert} training strategy to the multi-modal domain.
They accept both visual and textual token masked with a low masking rate as input and \yty{perform} the masked-token prediction task. 
(2) Discriminative methods:
These methods learn representations by contrastive learning or metric learning, \yty{by attracting paired video-text data and repelling unpaired one. }
\yty{For example,} Frozen \cite{bain2021frozen} leverages recent ViT\cite{dosovitskiy2020image} as visual encoder and \yty{can be flexibly trained} using both image and video datasets \yty{due to the sequential nature of transformer.}
OA-Trans \cite{wang2022object} adopts the bounding boxes and object tags to guide the training process. 
LocVTP \cite{cao2022locvtp} \yty{performs the fine-grained contrastive alignment to improve the localization performance of learned representation. }
Moreover, T2VLAD  \cite{wang2021t2vlad} and FCA \cite{han2021fine} also perform the ﬁne-grained interactions between video clips and text manually.
In contrast \yty{ to these complicated designs}, we \yty{present a simple and minimalist video-text pre-training framework based on the masked autoencoders and achieve performance far beyond current state-of-the-art methods}
\begin{figure*}
	\includegraphics[width=\linewidth]{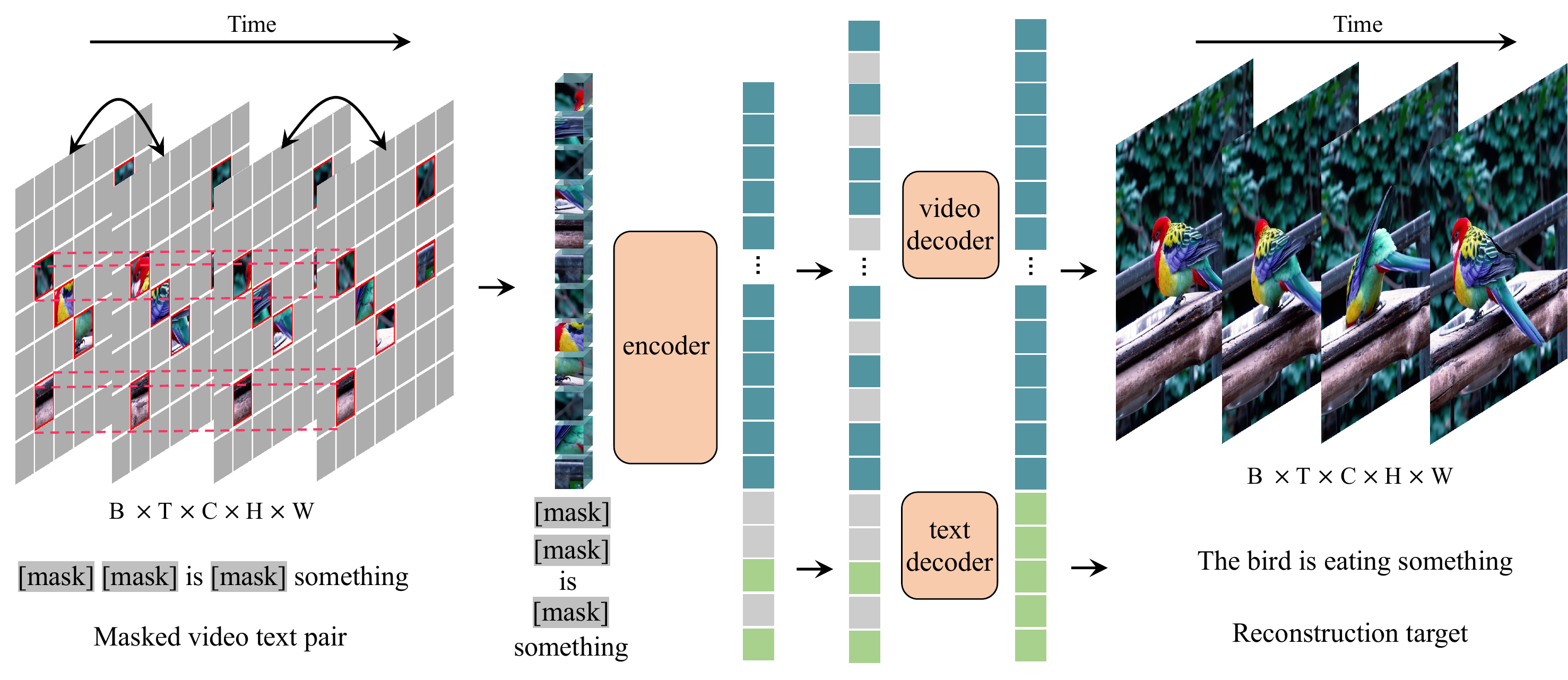}
	\caption{\textbf{Our SimVTP architecture.} By randomly masking out video tubes and text tokens with an extremely high mask ratio, SimVTP applies a unified encoder and two separate decoders to reconstruct the missing video and text. This unified encoder enables the model to learn cross correspondence by attention blocks in Transformer, benefiting the useful feature learning.}
	\label{fig:2}
	\vspace{-0.3cm}
\end{figure*}

\noindent\textbf{Masked Signal Modeling.}
\yty{Masked signal modeling, which reconstructs missing signal from corrupted one, is an effective way to learn representation for various signal modalities, e.g., text \cite{devlin2018bert, brown2020language}, image \cite{chen2020generative}  and video \cite{tong2022videomae}.  The most successful development of text representation learning is BERT \cite{devlin2018bert}, which significantly improves a broad range of downstream NLP tasks. For visual signals,}
early works \cite{pathak2016context} reconstruct the masking regions with context using convolutions, \yty{which do not achieves superior performance.  Recent methods are based on Transformer architectures, which is in line with NLP and achieves significant progress. } iGPT \cite{chen2020generative} \yty{treats image as a sequence of pixels, which are then fed into a transformer to train the model in an autoregressive way. However, because the number of pixels grows exponentially as the image size increases, iGPT \cite{chen2020generative} only conducts experiments on low-resolution images. }
\yty{BEiT \cite{bao2021beit} proposes to tokenize the image into visual tokens and reconstruct them from masked image patches.  }
MAE \cite{he2022masked} applies vision transformer as the backbone to reconstruct the masked area \yty{in pixels and obtains excellent performance. One of their key findings is using a high mask ratio to force the encoder to have holistic understanding of image. Both VideoMAE \cite{tong2022videomae} and \cite{feichtenhofer2022masked} extend MAE to video modality and observe that video is more redundant than image and therefore needs a higher mask ratio to obtain optimal performance. MaskFeat \cite{wei2022masked} instead predicts the video HOG features based on masked video tubes.  Inspired by these successes in both vision and NLP, we propose to unify video and text into a shared encoder and reconstruct the missing signal based on the masked video-text data. }

\section{Method}
\label{sec:Method}
\yty{SimVTP learns cross-modal representations by reconstructing the missing video cubes and text tokens conditioning on the corrupted input signals.  Our approach has a unified encoder which maps the concatenation of masked video tokens and text ones to a latent code, and two modality-specific decoder to reconstruct the original signal.  The learned encoder will be used by both video and text modality for various downstream tasks. Figure \ref{fig:2} illustrates the whole framework.}

\noindent\textbf{Video-text tokenizer.}
\yty{Before concatenating the video and text input into a single sequence, we need first map these two modalities to discrete tokens with same dimension. For video, we partition the video clip into a set of spatial-temporal cubes, where we apply a cube embedding layer \cite{arnab2021vivit, fan2021multiscale, tong2022videomae} to get the video tokens. For text, we map the input text into a sequence of text tokens using BERT\cite{devlin2018bert} tokenizer. }

\noindent\textbf{Video-text masking.}
\yty{We perform random masking on both video and text tokens following a uniform distribution as in \cite{tong2022videomae, devlin2018bert}. }
\yty{There are two potential strategies that can be used for video-text masking: 1) randomly masking the concatenation of video and text tokens as a whole using a single mask ratio, 2) randomly masking different modal tokens using separate mask ratio. We empirically observe that the second one performs better. This may be explained by that video and text have different information redundancy and therefore need different optimal mask ratio.}

\yty{It is well known that information-redundant signals, like images and videos, need a high masking ratio (e.g., 75\% in MAE \cite{he2022masked} and 90\% in VideoMAE \cite{tong2022videomae}) to construct a challenging self-supervisory task for good feature learning. Texts, which are  considered as highly semantic and information-dense,  usually use a low mask ratio (e.g., 15\% in BERT \cite{devlin2018bert}). However, this situation has changed when the text tokens are jointly reconstructed with paired video tokens. Given the fact that paired video and text convey the same information, the coupled video provides extra information redundancy for text recovering, making text reconstruction less challenging. We therefore need a higher mask ratio to reduce the information redundancy and facilitate useful feature learning. Our empirical experiment (See Figure \ref{fig:maskratio}) demonstrates that the optimal mask ratio of text is as high as 75\%. }

\noindent\textbf{SimVTP encoder.}
\yty{We use a shared Transformer \cite{attention-all-you-need} as the encoder of the concatenation of masked video and text tokens. Following \cite{tong2022videomae, he2022masked}, we only apply the encoder on visible video tokens, a small subset (e.g., 10\%) of total tokens, to save computation.  Since the length of text tokens is much smaller than videos (~18 text tokens vs 1568 video tokens), it does not really matter whether we operate unmasked tokens or full tokens in terms of computation.  Based on the empirical results in Section \ref{ablation}, we observe that employing the encoder on full text tokens, where the masked word tokens are replaced with a learnable vector, achieves the best performance. Therefore, we choose this setting as our default design. }

\noindent\textbf{SimVTP decoder.}
\yty{We reconstruct the missing video cubes and words using two separate decoders.  For video decoder, we use a lightweight Transformer that takes as input the full set of video tokens \cite{tong2022videomae}, where the masked ones are replaced by a shared learnable vector to indicate the presence of a missing video cube to be recovered.  To let the mask tokens have information of their locations in the video clip, we add positional embeddings to the full set of video tokens.  At the end of the video decoder, we apply a linear layer to project the representation to video cubes in pixel space. Since the text encoder already encode the mask tokens to imply the positional information of missing tokens, we simply use a linear layer for text decoder to predict the text token as in \cite{devlin2018bert}}

\noindent\textbf{Reconstruction target.}
\yty{For video reconstruction, we reshape the decoder's output to a reconstructed video clip and then apply the mean square error (MSE) loss between the reconstructed and the original video clips in pixel space \cite{tong2022videomae}. With regard to text reconstruction, we compute the cross entropy loss between the predicted and original text tokens \cite{devlin2018bert}. Note that we only compute the loss on masked tokens for both video and text. The masked signal modeling (MSM) loss $\mathcal{L}_{\mathrm{msm}}$ is the sum of these two losses, }

\noindent\textbf{Cross-modal training strategies.}
\yty{Besides training our model using masked signal modeling, we also apply two commonly used training strategies in video text pre-training community, i.e. video-text contrastive learning (VTC) and video-text matching (VTM). We empirically observe that these two schemes work pretty well on masked video and text tokens, which further boost the performance with a large margin.}

\yty{VTC aims to align the visual content and the corresponding text by pulling paired video-text data close and repelling unpaired ones apart. Specifically, given a batch of masked video-text samples, we separately compute their corresponding features using the shared encoder followed by a global average pooling layer, and then apply a symmetric cross-modal contrastive loss \cite{radford2021learning} $\mathcal{L}_{\mathrm{vtc}}$  on these features.}

\yty{VTM predicts whether a video-text pair is matched (positive) or not matched (negative).  Since the module of masked signal modeling has already computed the joint feature using the unified encoder for a batch of matched video-text pairs, we can reuse them and only need to construct the same number of unmatched video-text pairs randomly within the batch for feature computation. Concretely, for each video in a mini-batch, we randomly sample an unmatched text from the same batch to create an unpaired video-text input. We then append a linear layer followed by softmax as the classifier on these representations to predict a two-class probability. Finally, we calculate the cross entropy loss $\mathcal{L}_{\mathrm{vtm}}$ between the prediction and ground-truth label for optimization.}

The final pre-training loss of SimVTP is,
\begin{equation}
\mathcal{L}=\mathcal{L}_{\mathrm{msm}}+\mathcal{L}_{\mathrm{vtc}}+\mathcal{L}_{\mathrm{vtm}}
\label{eq:final_loss}
\end{equation}


\section{Experiment}
\label{sec:Exp}
\subsection{Experimental Setup}
We pre-train our SimVTP models on WebVid-2M \cite{bain2021frozen}, 
which includes 2.5M video-text pairs \footnote{We only obtain 2.3M video-text pairs due to some broken URLs.}.  
After pre-training our models, we evaluate them on the following downstream tasks:
1) \textbf{Text-to-Video-Retrieval} on MSRVTT \cite{xu2016msr}. 
For MSRVTT, 
\yty{We use the setting of 9K videos for training 1 K videos for testing.}
\yty{We use the shared encoder to compute the video and text representations separately and apply a cross-modal constrative loss on them for optimization. We follow the inference protocol in \cite{tong2022videomae} for video feature extraction during evaluation.}
2) \textbf{Video Question Answering} on TGIF \cite{li2016tgif}, MSRVTT\cite{xu2016msr}, LSMDC \cite{rohrbach2015dataset} and MSVD \cite{chen2011collecting}. \yty{There are two settings of VQA:  multiple-choice and open-ended. Multiple choice VQA only needs the model to pick an option out of the provided answers while open ended VQA requires the model to produce an natural language answer according to the input video and text query. 
For multiple choice VQA, we concatenate each candidate text with the given video, as well as the coupled query as the input of VQA model and then predict whether they are matched (positive) using a linear binary classification layer, which is the same as VTM head in pre-training.  Therefore, we load both the weights of SimVTP encoder and the VTM head from pre-trained model to initialize this VQA model.
For open ended VQA, we treat the task as a multi-class classification task by denoting all possible answers as class labels following the common practice \cite{lei2021less}. We use a two-layer MLP followed by a linear classification layer as the classifier. The input of this VQA model is the concatenation of the video and query tokens. 3) \textbf{Video Grounding} on ActivityNet Captions \cite{krishna2017dense}, Charades-STA \cite{gao2017tall} and TACos \cite{regneri2013grounding}. Following LocVTP \cite{cao2022locvtp}, we use the recent method 2D-TAN \cite{zhang2020learning} as our grounding baseline and replace the feature extractor with our SimVTP model to evaluate its performance. Further implementation details could be found in supplementary materials.}

\begin{table*}[t]
\centering
\begin{tabular}{llclllll}
	\toprule
	Method          & Vis Enc. Init      & Pre-trained Data & \#pairs & R@1  & R@5  & R@10 & MdR \\
	\midrule
	UniVL \cite{luo2020univl}           & \text { - }                  & HowTo100M        & 136M    & 21.2 & 49.6 & 63.1 & 6.0   \\
	ClipBERT \cite{lei2021less}       & \text { - }                  & COCO, VGen       & 5.6M    & 22.0   & 46.8 & 59.9 & 6.0   \\
	CE \cite{liu2019use}              & Multi-modal        & HowTo100M        & 136M    & 20.9 & 48.8 & 62.4 & 6.0   \\
	MMT \cite{gabeur2020multi}            & Multi-modal        & HowTo100M        & 136M    & 26.6 & 57.1 & 69.6 & 4.0   \\
	HIT \cite{liu2021hit}            & Multi-modal        & HowTo100M        & 136M    & 30.7 & 60.9 & 73.2 & 2.6 \\
	ActBERT \cite{zhu2020actbert}         & VisGenome          & HowTo100M        & 136M    & 16.3 & 42.8 & 56.9 & 10.0  \\
	SupportSet \cite{patrick2020support}     & IG65M, ImageNet    & HowTo100M        & 136M    & 30.1 & 58.5 & 69.3 & 3.0   \\
	HERO \cite{li2020hero}           & ImageNet, Kinetics & HowTo100M        & 136M    & 16.8 & 43.4 & 57.7 & \text { - }   \\
	NoiseEsti \cite{amrani2021noise} & ImageNet, Kinetics & HowTo100M        & 136M    & 17.4 & 41.6 & 53.6 & 8.0   \\
	DECEMBER \cite{tang2021decembert}       & ImageNet, Kinetics & HowTo100M        & 136M    & 17.5 & 44.3 & 58.6& 9.0 \\
	AVLnet \cite{rouditchenko2020avlnet}         & ImageNet, Kinetics & HowTo100M        & 136M    & 27.1 & 55.6 & 66.6 & 4.0   \\
	Frozen \cite{bain2021frozen}          & ImageNet           & WebVid2M+CC3M    & 5.5M    & 31.0   & 59.5 & 70.5 & 3.0   \\
	OATrans \cite{wang2022object}       & ImageNet           & WebVid2M+CC3M    & 5.5M    & 35.8 & 63.4 & 76.5 & 3.0   \\
	RegionLearner \cite{yan2021video}   & ImageNet           & WebVid2M+CC3M    & 5.5M    & 36.3 & 63.9 & 72.5 & 3.0   \\
	LocVTP \cite{cao2022locvtp}          & ImageNet           & WebVid2M+CC3M    & 5.5M    & 36.5 & 64.3 & 76.8 & 3.0   \\
	
	BFormer \cite{ge2022bridgeformer} & ImageNet           & WebVid2M+CC3M    & 5.5M    & 37.6 & 64.8 & 75.1 & 3.0   \\
	\textbf{SimVTP (Ours)}      & Kinetics           & WebVid2M         & 2.5M    & \textbf{53.6} &\textbf{81.9} & \textbf{90.7} & \textbf{1.0}   \\
	\toprule
	OATrans \cite{wang2022object}        & CLIP               & WebVid2M+CC3M    & 5.5M    & 40.9 & 70.4 & 80.3 & 2.0   \\
	BFormer \cite{ge2022bridgeformer}        & CLIP               & WebVid2M+CC3M   & 5.5M    & 44.9 & 71.9 & 80.0 & 2.0   \\
	VideoClip \cite{xu2021videoclip}       & CLIP               & HowTo100M        & 136M    & 30.9 & 55.4 & 66.8 & \text { - }   \\
	Clip4clip \cite{luo2022clip4clip}      & CLIP               & HowTo100M        & 136M    & 44.5 & 71.4 & 81.6 & 2.0   \\
	LocVTP \cite{cao2022locvtp}        & CLIP               & HowTo100M    & 136M    & 46.3 & 72.8 & 82.0 & 2.0   \\
	\textbf{SimVTP (Ours)}           & CLIP             & WebVid2M        & 2.5M   &  \textbf{61.0}    &   \textbf{87.7}   &  \textbf{94.5}    &  \textbf{1.0} \\
	\toprule
\end{tabular}
\caption{\textbf{Text-to-video retrieval results on MSRVTT.} Vis Enc. Init denotes datasets used for pre-training the video encoder. Higher R@k and lower MdR means better performance. The comparison results using CLIP initialization are shown at bottom.} \label{tb:retri-msrvtt}
\end{table*}

\begin{table*}[t]
\footnotesize
\centering
\renewcommand\arraystretch{1.1}
\resizebox{0.98\linewidth}{!}{
		\begin{tabular}{rcccc|cccc|cccc}
			\toprule
			\multirow{2}{*}{Method}  & \multicolumn{4}{c|}{ANet Captions} & \multicolumn{4}{c|}{Charades-STA} & \multicolumn{4}{c}{TACoS} \\
			~ & $R_1^{0.5}$ & $R_1^{0.7}$  & $R_5^{0.5}$ & $R_5^{0.7}$  & $R_1^{0.5}$ & $R_1^{0.7}$  & $R_5^{0.5}$ & $R_5^{0.7}$  & $R_1^{0.3}$ & $R_1^{0.5}$  & $R_5^{0.3}$ & $R_5^{0.5}$  \\
			\midrule
			
			VideoBERT\cite{sun2019videobert}   & 37.2 & 21.0 & 66.7 & 53.6 & 32.7 & 19.5 & 68.1 & 46.2 & 33.8 & 22.2 & 51.6 & 41.0   \\
			MIL-NCE~\cite{miech2020end}   & 41.8 & 24.5 & 73.5 & 57.7 & 37.0 & 21.2 & 74.3 & 50.4 & 35.1 & 23.5 & 53.7 & 42.5   \\
			UniVL~\cite{luo2020univl}   & 42.2 & 25.4 & 75.3 & 60.5 & 38.2 & 22.7 & 77.2 & 51.4 & 35.7 & 23.7 & 55.8 & 43.7  \\
			December~\cite{tang2021decembert} & 43.0 & 25.1 & 76.0 & 60.2 & 37.2 & 21.6 & 78.3 & 50.6 & 34.8 & 22.9 & 55.1 & 43.9 \\
			Frozen~\cite{bain2021frozen} & 43.3 & 25.8 & 75.8 & 59.3 & 38.8 & 22.9 & 77.6 & 50.3 & 35.7 & 23.5 & 54.4 & 43.7 \\
			OA-Trans\cite{wang2022object} & 43.6 & 25.9 & 76.5 & 60.2 & 39.2 & 22.6 & 78.5 & 50.8 & 35.2 & 22.5 & 53.4 & 42.6 \\
			LocVTP\cite{cao2022locvtp} & 46.1 & 27.6 & 78.9 & 63.7 & {41.2} & {24.8} & {81.3} & {53.5} & {39.6}& {27.8} & {60.4} & {47.9} \\ 
			\textbf{SimVTP (Ours)} & \textbf{49.2} & \textbf{29.4} & \textbf{80} & \textbf{65.1} & \textbf{44.7} & \textbf{26.3} & \textbf{83.7}   & \textbf{55.1} & \textbf{43.1} & \textbf{30.3} & \textbf{63.1} & \textbf{49.6}  \\ 
			\bottomrule
		\end{tabular}
	}
	\caption{\textbf{Video grounding results on various datasets.} $R_n^{m}$ means the percentage of at least one successful retrieved moments among the top-n, where moments whose IOU with ground-truth are larger than m are regarded as success. Higher $R_n^{m}$ indicates better performance}\label{tb:vg}
	\vspace{-0.3cm}
\end{table*}

\begin{table}[]
	\centering
	\setlength\tabcolsep{5.pt}
	\resizebox{0.98\linewidth}{!}{
		\begin{tabularx}{1.06\linewidth}{p{2.1cm}p{0.4cm}p{0.4cm}p{0.4cm}p{0.45cm}p{0.4cm}p{0.95cm}p{0.4cm}}
			\toprule
			\multirow{2}{*}{Method} & \multicolumn{3}{c}{TGIF}    & \multicolumn{2}{c}{MSRVTT} & LSMDC & MSVD \\
			& A & T & F & MC           & OE          & MC    & OE  \\
			\toprule
			JSFusion \cite{yu2018joint} & \text { - } & \text { - } & \text { - } & 83.4 & \text { - } & 73.5  & \text { - } \\
			HTGS \cite{fan2019heterogeneous} & 79.3 & 77.8 & 53.8 & \text { - } & 33 & \text { - }  & \text { - } \\
			ClipBert \cite{lei2021less}               & 82.8   & 87.5       & 59.4  & 88.2         & 37.4        & \text { - }     & \text { - }    \\
			JuskAsk \cite{yang2021just}                & \text { - }      & \text { - }          & \text { - }     & \text { - }            & 41.5        & \text { - }     & 46.3 \\
			ALPRO \cite{li2021align}                  & \text { - }      & \text { - }          & \text { - }     & \text { - }            & 42.1        & \text { - }     & 45.9 \\
			VIOLET \cite{fu2021violet} & 92.5      & 95.7       & 68.9     & 91.9            & 43.9        & 82.8     & 47.9 \\
			MERLOT \cite{zellers2021merlot}  & 94.0      & 96.2       & 69.5     & 90.9            & 43.1        & 81.7     & \text { - } \\
			\textbf{SimVTP}  & \textbf{94.4}   & \textbf{96.9}       & \textbf{70.2}  & \textbf{93.6}        & \textbf{44.7}        &\textbf{83.7}  & \textbf{48.9} \\
			\bottomrule
	\end{tabularx}}
	\caption{\textbf{Video question answering results.} A: Action, T: Transition, F: FrameQA, MC: Multiple Choice, OE: Open Ended}\label{tb:vqa}
	\vspace{-0.2cm}
\end{table}

\subsection{Comparing to State-of-the-art}

\noindent\textbf{Text-to-video retrieval.}
\yty{Table \ref{tb:retri-msrvtt} presents the text-to-video retrieval results on MSR-VTT \cite{xu2016msr}. Our SimVTP overwhelmingly outperforms all previous methods by a large margin, despite being pre-trained only on WebVid-2M \cite{bain2021frozen}.  With CLIP weights trained on 400 million image-text pairs as encoder initialization, our SimVTP achieves a significant high performance of 61.0\% R@1.  Particularly,  our SimVTP initialized with VideoMAE \cite{tong2022videomae} weights trained on Kinetics-400 \cite{kay2017kinetics} even surpasses most methods that use CLIP weights as initialization. \yty{More comparison results on MSVD \cite{chen2011collecting}, DiDemo \cite{anne2017localizing} and LSMDC \cite{rohrbach2015dataset} can be found in supplementary materials. }}

\noindent\textbf{Video question answering.} 
\yty{Table \ref{tb:vqa} reports the comparison results of VQA on four datasets under different settings.} Our SimVTP beats all previous algorithms, though using much less pre-training data. Moreover, our SimVTP even outperforms JuskAsk \cite{yang2021just}, which is dedicatedly designed for VQA and trained on orders of magnitude larger datasets (e.g., HTVQA69M , YTT180M), on both MSRVTT-QA and MSVD-QA settings.

\noindent\textbf{Video Grounding.} 
\yty{Table \ref{tb:vg} lists the comparison results of video grounding on three datasets. Our SimVTP achieves the best performance compared with recent state-of-the-art methods. Particularly, our SimVTP surpasses recent video-text pre-training method LocVTP \cite{cao2022locvtp}, which is specially designed for video grounding, with a large margin, demonstrating the effectiveness of our method. What's more, our SimVTP only uses WebVid-2M for training while other methods like Frozen \cite{bain2021frozen}, OA-Trans \cite{wang2022object} and LocVTP \cite{cao2022locvtp} are all pre-trained on both WebVid-2M \cite{bain2021frozen} and CC3M \cite{sharma2018conceptual}, which further indicates the data efficiency of our algorithm.}

\begin{figure}[t]
	\includegraphics[width=8cm]{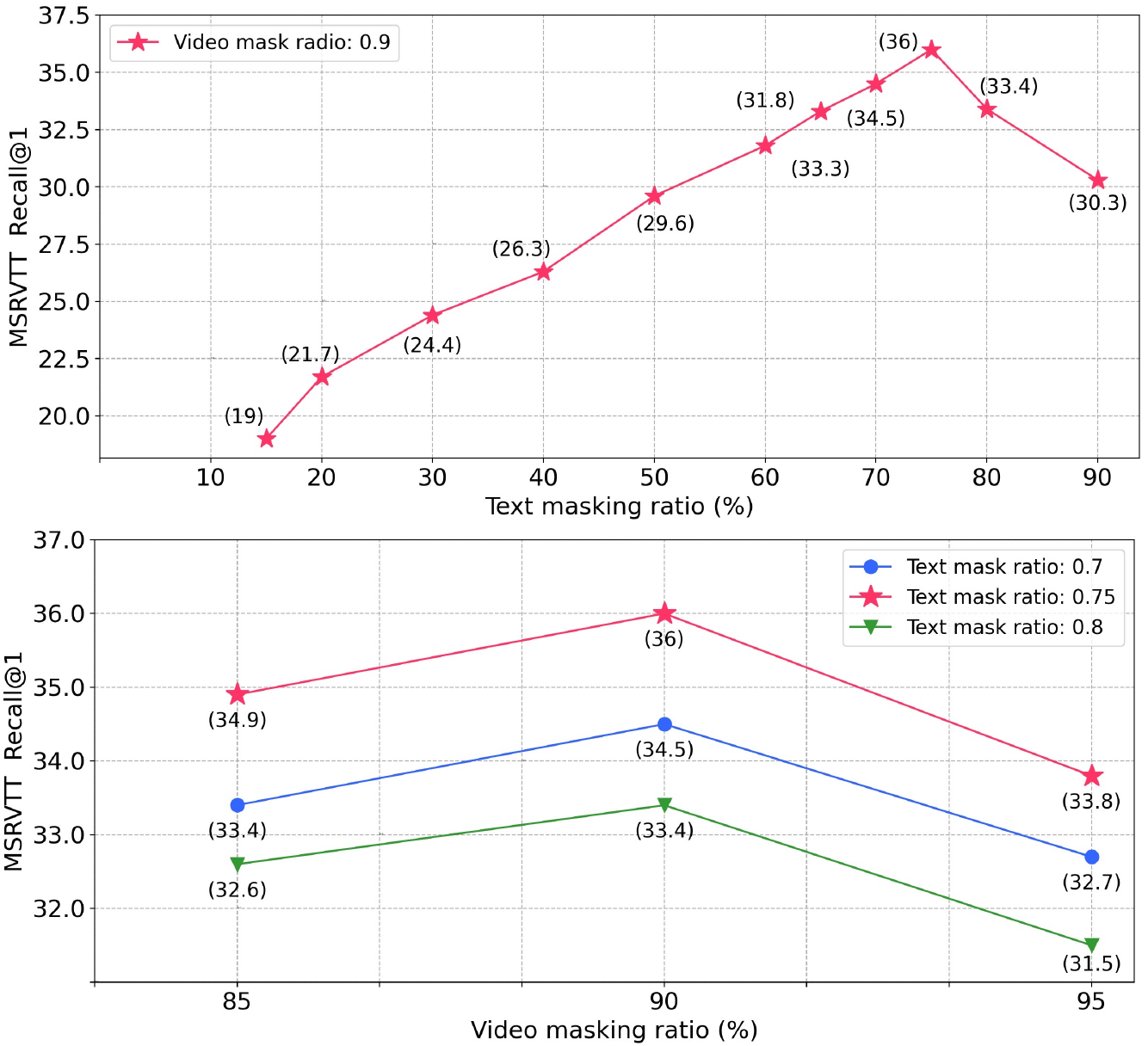}
	\caption{\textbf{The effect of mask ratio.} The performance is improved gradually as the text mask ratio increases and reaches the optimal at 75\% (top). Video mask ratio works best at 90\% under different text mask ratios (bottom). }\label{fig:maskratio}
	\vspace{-0.5cm}
\end{figure}

\definecolor{Gray}{gray}{0.85}
\begin{table*}
	\centering
	\begin{subtable}[c]{0.3\textwidth}
		\centering
		\begin{tabularx}{0.9\linewidth}{p{2.cm}p{0.6cm}p{0.5cm}}
			\toprule
			video blocks & R@1 & R@5 \\
			\midrule
			1                & 33.9 & 63.7 \\
			2                & 35.2 & 65.1\\
			4                 &  \cellcolor{Gray}\textbf{36.0}  & \cellcolor{Gray}\textbf{66.8}\\
			8                & 34.4 & 64.4 \\
			\bottomrule
		\end{tabularx}
		\vspace{0.3cm}
		\subcaption{\textbf{Video decoder depth.} Increasing the depth improves the result and start to saturates at 4.  }\label{tb1:a}
	\end{subtable} \quad \quad
	\begin{subtable}[c]{0.3\textwidth}
		\centering
		\begin{tabularx}{0.9\linewidth}{p{2.cm}p{0.6cm}p{0.5cm}}
			\toprule
			text blocks  & R@1  & R@5 \\
			\midrule
			0				& \cellcolor{Gray}\textbf{36.0} & \cellcolor{Gray}\textbf{66.8}\\
			1              & 35.7 & 65.8\\
			2                & 36.2 & 67.1\\
			4                 & 36.0 & 66.9  \\
			\bottomrule
		\end{tabularx}
		\vspace{0.3cm}
		\subcaption{\textbf{Text decoder depth.} Deeper text decoder achieves similar result with shallow one.}\label{tb1:b}
	\end{subtable}\qquad
	\begin{subtable}[c]{0.31\textwidth}
		\centering
		\begin{tabularx}{0.9\linewidth}{p{2.5 cm}p{0.6cm}p{0.5cm}}
			\toprule
			pretrain weights   & R@1  &R@5\\
			\midrule
			\textit{from scratch} & 27.4  & 60.0 \\
			BERT  & 29.9  & 62.9 \\
			ImgNet21K-Sup  &32.1   & 64.1 \\
			ImgNet1K-MAE &34.7  & 64.8 \\
			K400-VidMAE  & \cellcolor{Gray}\textbf{36.0} & \cellcolor{Gray}\textbf{66.8}\\
			\bottomrule
		\end{tabularx}
		\subcaption{\textbf{Unified encoder initialization.} VideoMAE weights pre-trained on K-400 performs best.}\label{tb1:c}
	\end{subtable}\qquad
	\begin{subtable}[c]{0.3\textwidth}
		\centering
		\vspace{0.3cm}
		\begin{tabularx}{0.9\linewidth}{p{2.cm}p{0.6cm}p{0.5cm}}
			\toprule
			case     . & R@1  &FLOPs\\
			\midrule
			text  w/ [M]       &  \cellcolor{Gray}\textbf{36.0}  & \cellcolor{Gray}\textbf{1.1×}\\
			text  w/o [M]       & 35.1  & 1×\\
			\bottomrule
		\end{tabularx}
		\vspace{0.cm}
		\subcaption{\textbf{Text mask token.} Using text with mask token as input improves the performance.}\label{tb1:d}
	\end{subtable}\qquad
	\begin{subtable}[c]{0.3\textwidth}
		\centering
		\vspace{0.15cm}
		\begin{tabularx}{0.9\linewidth}{p{2.0cm}p{0.6cm}p{0.5cm}}
			\toprule
			case     . & R@1  & R@5 \\
			\midrule
			dual decoder   & \cellcolor{Gray}\textbf{36.0}  &\cellcolor{Gray}\textbf{66.8}\\
			share decoder   & 34.1  & 64.0\\
			\bottomrule
		\end{tabularx}
		\vspace{0.cm}
		\subcaption{\textbf{Decoder type.} Dual decoder achieves better results compared with shared one.}\label{tb1:e}
	\end{subtable}\qquad
	\begin{subtable}[c]{0.31\textwidth}
		\centering
		\vspace{0.2cm}
		\begin{tabularx}{0.9\linewidth}{p{2.5 cm}p{0.6cm}p{0.5cm}}
			\toprule
			masking strategy & R@1 & R@5 \\
			\midrule
			MAE-like              & 35.6 & 66.1\\
			BERT-like                & \cellcolor{Gray}\textbf{36.0} & \cellcolor{Gray}\textbf{66.8}\\
			\bottomrule
		\end{tabularx}
		\subcaption{\textbf{Text masking strategies.} Bert-like text masking strategy improves the performance.}\label{tb1:f}
	\end{subtable}
	\caption{\textbf{SimVTP ablation experiments} with 16-frame ViT-B on MSR-VTT. All these experiments are conducted with only MSM for 100 epochs on WebVid-240K. The video mask ratio and text mask ratio are 90\% and 75\%. Default settings are  in \colorbox{Gray} {gray}.} 
\end{table*}

\begin{table}[]
	\centering
	\begin{tabular}{lllll}
		\toprule
		MSM & VTC & VTM  &   R@1       &   R@5        \\
		\midrule
		\textcolor{darkergreen}{\Checkmark}      &   \XSolidBrush    & \XSolidBrush           & 36.0          & 66.8         \\
		\XSolidBrush    & \textcolor{darkergreen}{\Checkmark}      & \XSolidBrush              & 36.8    & 67.0             \\
		\XSolidBrush    & \XSolidBrush      & \textcolor{darkergreen}{\Checkmark}             &  24.6    & 54.5             \\
		\textcolor{darkergreen}{\Checkmark}      & \textcolor{darkergreen}{\Checkmark}        & \XSolidBrush              & 42.3    &   69.3              \\
		\textcolor{darkergreen}{\Checkmark}      & \XSolidBrush      & \textcolor{darkergreen}{\Checkmark}            &   39.8     &   67.4          \\
		\XSolidBrush    & \textcolor{darkergreen}{\Checkmark}       & \textcolor{darkergreen}{\Checkmark}             &  41.1     &   67.9           \\
		\textcolor{darkergreen}{\Checkmark}      & \textcolor{darkergreen}{\Checkmark}        & \textcolor{darkergreen}{\Checkmark}             &  43.8     & 70.9          \\
		\bottomrule
	\end{tabular}
	\caption{\textbf{SimVTP with different training strategies.} i.e. video-text contrastive learning (VTC) and video-text matching (VTM).}\label{tb:loss}
	\vspace{-0.5cm}
\end{table}

\subsection{Ablation Studies}\label{ablation}
\yty{We conduct extensive ablation experiments to explore the optimal design choice for our SimVTP. To save training time, we randomly sample a subset of WebVid-2M \cite{bain2021frozen} and obtain 240K video-text samples, which we called WebVid-240K, for ablation experiments training. All models are pre-trained for 100 epochs. Unless stated otherwise, we perform all the ablation studies with the default 16-frame ViT-B backbone and report the video retrieval performance on MSR-VTT \cite{xu2016msr}. Note that all models are trained only by MSM loss except for ablations on cross-modal training strategies.}

\noindent\textbf{Masking ratio}.
\yty{Figure \ref{fig:maskratio} presents the influence of masking ratio for both video and text tokens.  Under the video mask ratio of 90\%, the optimal text mask ratio is as high as 75\% as shown in Figure \ref{fig:maskratio} (top). This is in contrast to BERT \cite{devlin2018bert}, which usually uses a mask ratio of 15\%.  Higher text mask ratios, like 80\% and 90\% can still obtain decent performance while lower text mask ratio, like 15\% and 20\% perform poorly. This behavior indicates that the coupled video provides extra information redundancy for text and thus higher mask ratios are needed to reduce redundancy and build a challenging self-supervised task for useful feature learning. By fixing the text mask ratio to a relative high mask ratio, e.g., 70\%, 75\% and 80\%, Figure \ref{fig:maskratio} (bottom) varies the video mask ratio and demonstrates that the optimal video mask ratio is 90\%, which is inline with \cite{tong2022videomae}}

\noindent\textbf{Decoder design}. 
\yty{There are two possible ways to design our SimVTP decoder: 1) dual decoder, which uses two separate decoders for video and text reconstruction respectively. 2) share decoder, which applies a share decoder with two separate head for video and text reconstruction respectively.  Table \ref{tb1:e} shows that dual decoder achieves better performance. This can be explained by the large gap between reconstructing video cubes and text tokens, making it hard to reconstruct both modalities well using a shared decoder. Since only visible video tokens are fed into the encoder, the video decoder needs several Transformer blocks to propagate information between visible and mask tokens for reconstruction. The optimal depth (number of Transformer blocks) for video decoder is 4, as shown in Table \ref{tb1:a}. However, this is not the case for text decoder. Because both visible and mask text tokens are input to the encoder, a heavy decoder for text reconstruction becomes unnecessary \cite{devlin2018bert, xie2022simmim}, . As demonstrated in Table \ref{tb1:b}, a simple linear layer works pretty well compared with other heavier design. The video decoder width is set to half channel of the encoder (e.g., 384-d for ViT-B), following \cite{tong2022videomae}}

\noindent\textbf{Mask token.}
\yty{We follow the design choice of MAE \cite{he2022masked} and VideoMAE \cite{tong2022videomae} that skips the video mask token [M] in the encoder and then insert it in the decoder. This design greatly reduces the training time and GPU memory consumption. However, we observe that performing this strategy on text hurts the performance. Table \ref{tb1:d} shows that using text tokens with mask token [M] filling the missing positions as input of encoder improves the performance. The length of text tokens is relatively short ($\sim$18 words), which is in contrast to video tokens (1568 cubes). Therefore skipping the text mask token in the encoder does not reduce too much training computation. We choose to input text tokens with mask token [M] for encoder by default due to the performance gain.}

\begin{figure*}[t]
	\centering
	\includegraphics[width=\textwidth]{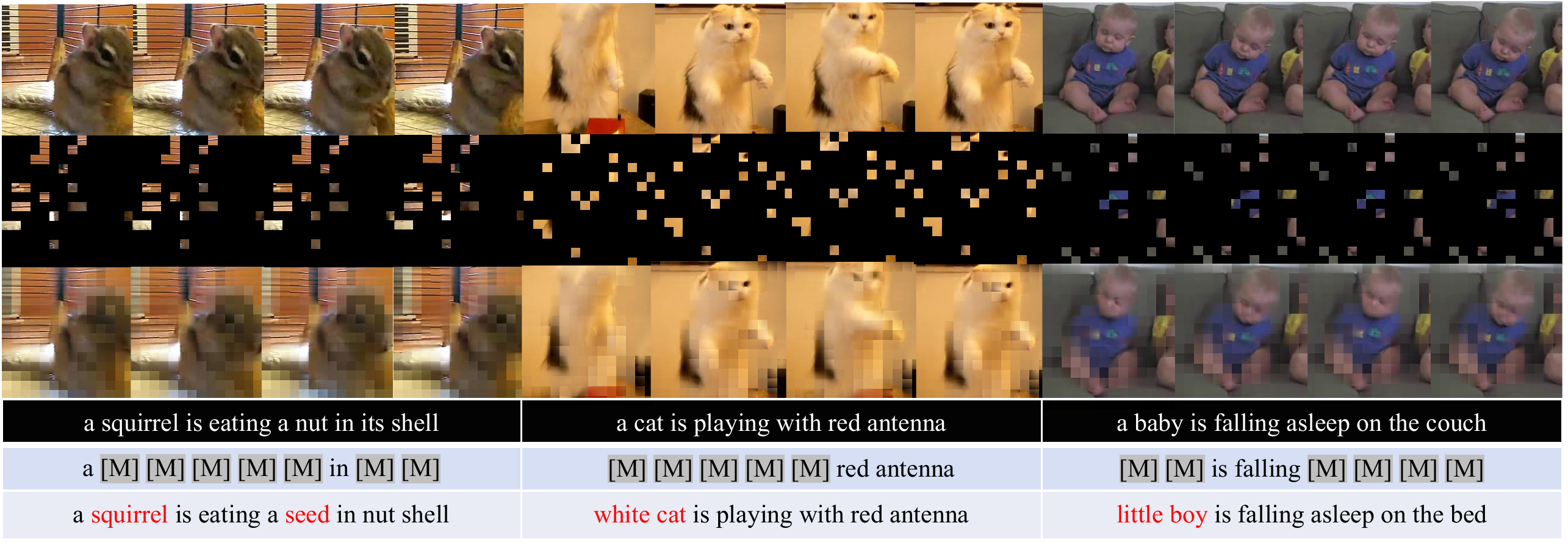}
	\caption{\textbf{Vido-text reconstruction} on WebVid-2M using a SimVTP pre-trained with video mask ratio of 90\% and text mask ratio of 75\%. From top to bottom, we show original frames, masked frames, reconstructed frames, original text, masked text and reconstructed text.}\label{fig:rec}
\end{figure*}

\begin{figure*}[t]
	\centering
	\includegraphics[width=\textwidth]{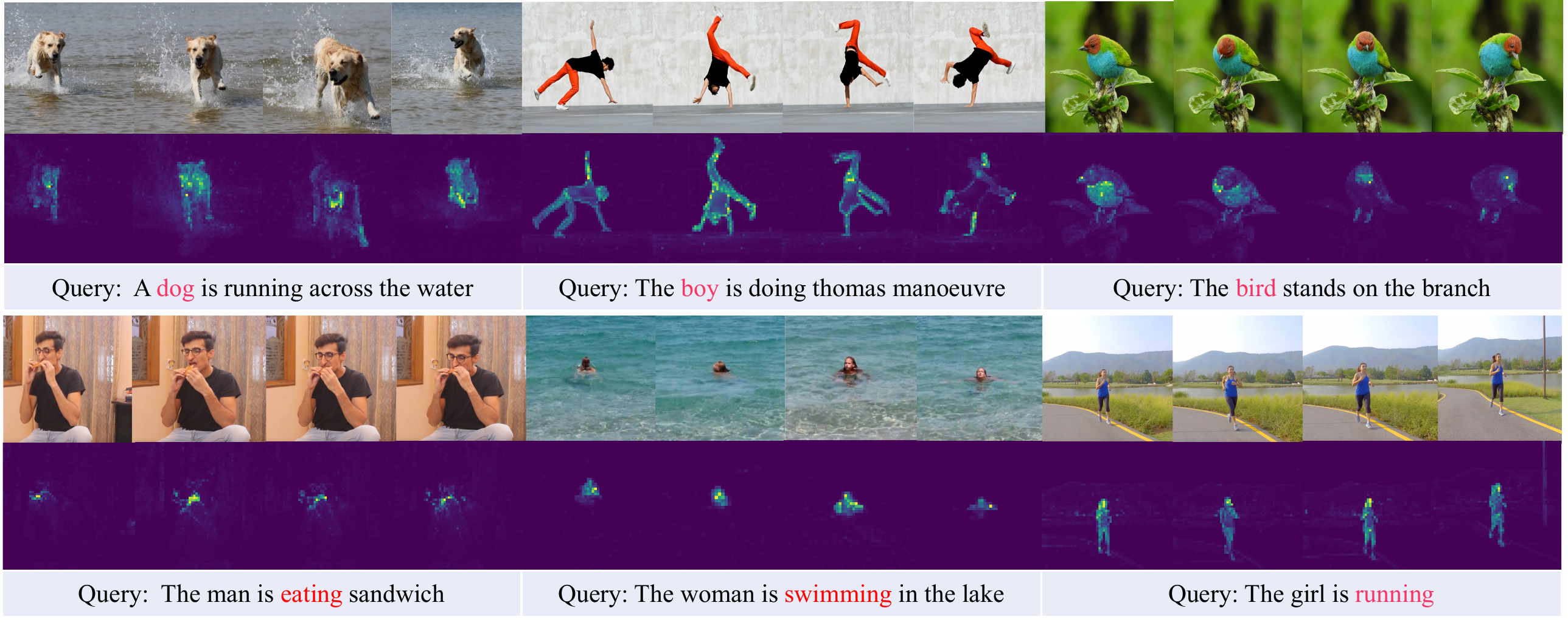}
	\caption{\textbf{Cross attention weight} from the unified encoder learned by our SimVTP. The reference word token is marked in red. The words in the top row are nouns and the ones in the bottom row are verbs. Both types of words can attend the visual content well.}
	\label{fig:attention}
\end{figure*}

\noindent\textbf{Masking strategies}. For video tokens, we follow the masking strategies of VideoMAE \cite{tong2022videomae}, which has been demonstrated to be effective for video modality.  For text tokens, we try two strategies: 1) MAE-like strategy simply replaces the masked word with a mask token [M]; 2) BERT-like strategy replaces the masked word with a mask token [M], a random word or the same word with the chances of 80\%, 10\% and 10\% respectively.  Table \ref{tb1:f} demonstrates that BERT-like strategy works better.

\noindent\textbf{Encoder initialization}.
\yty{Encoder initialization has a great impact on multi-modal pre-training. In Table \ref{tb1:c}, we compare different encoder weight initialization for SimVTP. We observe that VideoMAE weights pre-trained on Kinetics-400 work best. 
We test the performance of using ImageMAE weights pre-trained on ImageNet-1K and find that it even outperforms the performance of supervised weights pre-trained on ImageNet-21K, demonstrating the effectiveness of masked autaoencoder pre-training. \ytyn{We further use BERT weights to initialize our unified encoder for pre-training and observe degraded performance, indicating that text transformer weights are less effective for video-text tasks compared with vision one. Training without any initialization achieves the worst performance, demonstrating the importance of encoder initialization.}}

\noindent\textbf{Cross-modal training strategies}. 
\yty{Aside from training our model using masked signal modeling (MSM), we also apply two extra commonly used cross modal training strategies, i.e. video-text contrastive learning (VTC) and video-text matching (VTM), on our model. Table \ref{tb:loss} presents the comparison results. Both VTC and VTM boost the performance greatly. Interestingly, we observe that only performing VTC on masked video-text tokens could obtain a similar performance compared with MSM. This result is counter-intuitive at the first glance, since we all know that mainstream contrastive video-text pretraining methods do not achieve such good performance.  For instance, Frozen \cite{bain2021frozen} only achieves 31.0\%R@1 by using both CC3M and WebVid-2M, which is 5.5 M data pairs, while our model only uses 1/10 of WebVid-2M, i.e., 240K data pairs. We further conduct another experiment that performs VTC on original video-text pairs \textit{without masking out any tokens} and obtains 25.7\%R@1, which is much lower than our VTC pre-trained on masked tokens (36.8\%R@1). This demonstrates the importance of token masking, which thus could be regarded as an effective data augmentation for contrastive learning. By introducing these two training schemes, our SimVTP achieves 43.7\%R@1 on MSR-VTT with only 240K video-text pairs as training data, which outperforms most recent state-of-the-art methods pre-trained on both WebVid-2M \cite{bain2021frozen} and CC3M \cite{sharma2018conceptual} by a large margin.}

\subsection{Visualization}
\label{sec:Vis}
\yty{We visualize the video-text reconstruction examples in Figure \ref{fig:rec} and observe very interesting reconstructed texts. Though not the exact same as original texts, the reconstructed texts are plausible and in harmony with the video content. Sometimes, they are even more accurate than original texts, like the 'white cat'  and 'little boy' in the second and third columns of Figure \ref{fig:rec}. We also visualize the cross attention weights of text tokens on video frames in Figure \ref{fig:attention}. 
We can see that SimVTP learns the cross-modal correspondence well. \ytyn{The word tokens including both noun (top) and verb (bottom) in Figure \ref{fig:attention}} could decently attend corresponding video frame patches. For instance, the action word 'eating' corresponds to the mouth of the man accurately.}

\section{Conclusion}
\label{sec:Con}
In this paper, we propose a simple video-ext pre-training framework (SimVTP) with masked autoencoder, which utilizes a unified autoencoder to reconstruct the masked signal of one modality with the help from another modality. The extremely high mask ratios for both video  and text are leveraged to encourage the model to learn useful representations.  \textcolor{black}{Equipping SimVTP with video-text contrastive learning (VTC) and video-text matching (VTM), could further improve the transferable performance.} Moreover, SimVTP is data-efficient. Using only WebVid-2M as training data, SimVTP achieves new state-of-the-art results on a broad range of downstream tasks, including video retrieval, video question answering and video grounding.

\balance
{\small
	\bibliographystyle{ieee_fullname}
	\bibliography{egbib}
}
\clearpage

	
\end{document}